# An Efficient Secure Multimodal Biometric Fusion Using Palmprint and Face Image

Nageshkumar.M, Mahesh.PK and M.N. Shanmukha Swamy

[1] *Department of Electronics and Communication,*
*J.S.S. research foundation, Mysore University, Mysore-6*
nageshkumar79m@gmail.com

[2] *Department of Electronics and Communication,*
*J.S.S. research foundation, Mysore University, Mysore-6*
mahesh24pk@gmail.com

[3] *Department of Electronics and Communication,*
*J.S.S. research foundation, Mysore University, Mysore-6*
mnsjce@gmail.com

**Abstract**
Biometrics based personal identification is regarded as an effective method for automatically recognizing, with a high confidence a person's identity. A multimodal biometric systems consolidate the evidence presented by multiple biometric sources and typically better recognition performance compare to system based on a single biometric modality. This paper proposes an authentication method for a multimodal biometric system identification using two traits i.e. face and palmprint. The proposed system is designed for application where the training data contains a face and palmprint. Integrating the palmprint and face features increases robustness of the person authentication. The final decision is made by fusion at matching score level architecture in which features vectors are created independently for query measures and are then compared to the enrolment template, which are stored during database preparation. Multimodal biometric system is developed through fusion of face and palmprint recognition.

## 1 INTRODUCTION

A multimodal biometric authentication, which identifies an individual person using physiological and/or behavioral characteristics, such as face, fingerprints, hand geometry, iris, retina, vein and speech is one of the most attractive and effective methods. These methods are more reliable and capable than knowledge-based (e.g. Password) or token-based (e.g. Key) techniques. Since biometric features are hardly stolen or forgotten.

However, a single biometric feature sometimes fails to be exact enough for verifying the identity of a person. By combining multiple modalities enhanced performance reliability could be achieved. Due to its promising applications as well as the theoretical challenges, multimodal biometric has drawn more and more attention in recent years [1]. Face and palmprint multimodal biometrics are advantageous due to the use of non-invasive and low-cost image acquisition. In this method we can easily acquire face and palmprint images using two touchless sensors simultaneously. Existing studies in this approach [2, 3] employ holistic features for face representation and results are shown with small data set that was reported.

Multimodal system also provides anti-spooling measures by making it difficult for an intruder to spool multiple biometric traits simultaneously. However, an integration scheme is required to fuse the information presented by the individual modalities.

This paper presents a novel fusion strategy for personal identification using face and palmprint biometrics [8] at the features level fusion Scheme. The proposed paper shows that integration of face and palmprint biometrics can achieve higher performance that may not be possible using a single biometric indicator alone. This paper presents a new method called canonical form based on PCA, which gives better performance and better accuracy for both traits (face & palmprint).

The rest of this paper is organized as fallows. Section 2 presents the system structure, which is used to increase the performance of individual biometric trait; multiple

IJCSI

classifiers are combined using matching scores. Section 3 presents feature extraction using canonical form based on PCA. Section 4, the individual traits are fused at matching score level using sum of score techniques. Finally, the experimental results are given in section 5. Conclusions are given in the last section.

## 2  SYSTEM STRUCTURE

The multimodal biometrics system is developed using two traits (face & palmprint) as shown in the figure1. For both, face & palmprint recognition the paper proposes a new approach called canonical form based on PCA method for feature extraction. The matching score for each trait is calculated by using Euclidean distance. The modules based on individual traits returns an integer value after matching the templates and query feature vectors. The final score is generated by using the sum of score technique at fusion level, which is then passed to the decision module. The final decision is made by comparing the final score with a threshold value at the decision module.

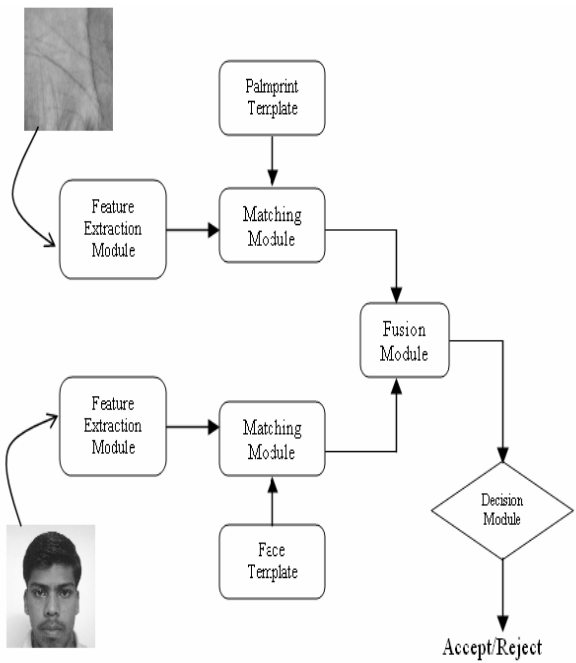

Figure1. Block diagram of face and palmprint multimodal biometric system

## 3  FEATURE EXTRACTION USING CANONICAL FORM BASED ON PCA APPROACH

The "Eigenface" or "Eigenpalm" method proposed by Turk and Pentland [5] [6] is based on Karhunen-Loeve Expression and is motivated by the earlier work of Sirovitch and Kirby [7][8] for efficiently representing picture of images. The Eigen method presented by Turk and Pentland finds the principal components (Karhunen-Loeve Expression) of the image distribution or the eigenvectors of the covariance matrix of the set of images. These eigenvectors can be thought as set of features, which together characterized between images.

Let a image $I(x, y)$ be a two dimensional array of intensity values or a vector of dimension $n$. Let the training set of images be $I_1, I_2, I_3, \ldots I_n$. The average image of the set is defined by

$$\Psi = \frac{1}{N}\sum_{i=1}^{n} I\,i \qquad (1)$$

Each image differed from the average by the vector. This $\phi_I = I\,i - \Psi$ set of very large vectors is subjected to principal component analysis which seeks a set of $K$ orthonormal vectors $V_k, K=1,\ldots, K$ and their associated eigenvalues $\lambda k$ which best describe the distribution of data. The vectors $V_k$ and scalars $\lambda k$ are the eigenvectors and eigenvalues of the covariance matrix:

$$C = \frac{1}{N}\sum_{i=1}^{N}\phi_i\,\phi_i^T = A\,A^T \qquad (2)$$

Where the matrix $A = [\phi_1, \phi_2 \ldots \ldots \phi_N]$ finding the eigenvectors of matrix $Cnxn$ is computationally intensive. However, the eigenvectors of $C$ can determine by first finding the eigenvectors of much smaller matrix of size $NxN$ and taking a linear combination of the resulting vectors [6].

The canonical method proposed in this paper is based on Eigen values and Eigen vectors. These Eigen valves can be thought a set of features which together characterized between images.

Let $Q$ be a quadratic form given by

$$Q = C^T I\,C = \sum_{i=1}^{n}\sum_{j=1}^{n} a_{ij}\,c_i\,c_j \qquad (3)$$

IJCSI

Therefore "n" set of eigen vectors corresponding "n" eigen values.

Let $\hat{P}$ be the normalized modal matrix of I, the diagonal matrix is given by

$$\hat{P}^{-1} I \hat{P} = D$$

Where    $I = \hat{P} D \hat{P}^{-1}$    (4)

Then
$$Q = C^T I C = C^T \hat{P} D \hat{P}^{-1} C = (C^T \hat{P})(D)(\hat{P}^{-1} C)$$    (5)

The above equation is known as a canonical form or sum of squares form or principal axes form.

The following steps are considered for the feature extraction:

(1) Select the text image for the input
(2) Pre-process the image (only for palm image)
(3) Determine the eigen values and eigen vectors of the image
(4) Use the canonical for the feature extraction.

**3.1 EUCLIDEAN DISTANCE:** Let an arbitrary instance X be described by the feature vector

$X = [a_1(x), a_2(x).........a_n(x)]$ Where $a_r(x)$ denotes the value of the $r^{th}$ attribute of instance x. Then the distance between two instances $x_i$ and $x_j$ is defined to be $d(x_i, x_j)$;

$$d(x_i, x_j) = \sqrt{\sum_{r=1}^{n}(a_r(X_i) - a_r(X_j))^2}$$    (6)

## 4  FUSION

The different biometric system can be integrated to improve the performance of the verification system. The following steps are performed for fusion:

(1) Given the query image as input, features are extracted by a individual recognition.
(2) The weights $\alpha$ and $\beta$ are calculated.
(3) Finally the sum of score technique is applied for combining the matching score of two traits i.e. face and palmprint. Thus the final score $MN_{FINAL}$ is given by

$$MS_{FINAL} = \frac{1}{2}(\alpha * MS_{Face} + \beta * MS_{Palm})$$    (7)

Where $\alpha$ and $\beta$ are the weights assigned to both the traits. The final matching score ($MS_{FINAL}$) is compared against a certain threshold value to recognize the person as genuine or an impostor.

## 5  EXPERIMENTAL RESULTS

We evaluate the proposed multimodal system on a data set including 720 pairs of images from 120 subjects. The training database contains a face & palmprint images for each individual for each subject.

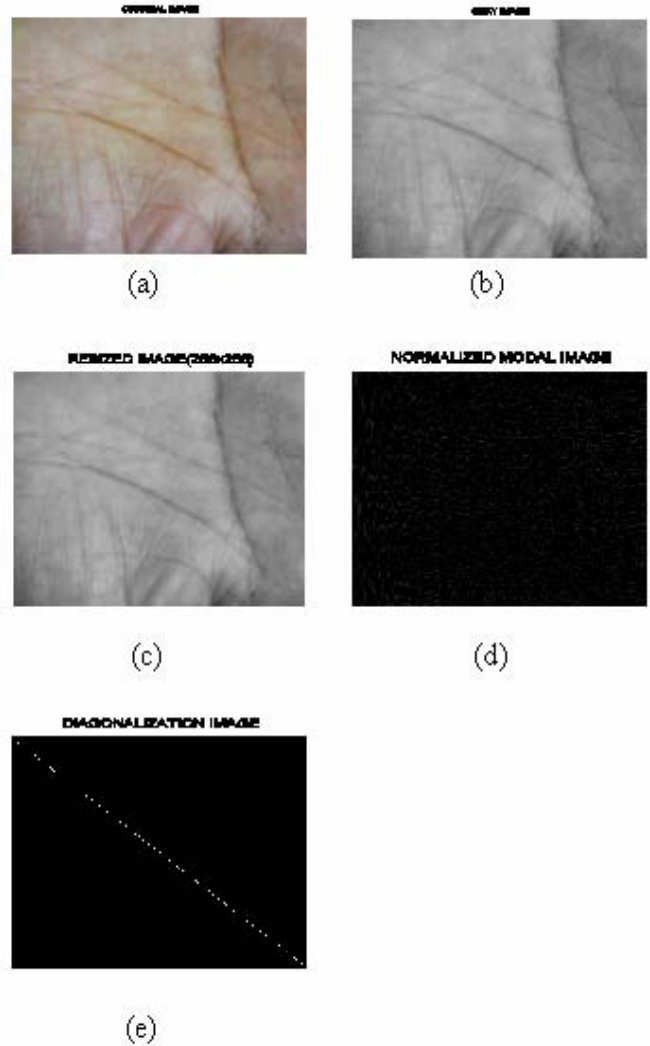

(a)    (b)

(c)    (d)

(e)

Fig.2. Canonical form based Palm images.
(a) Original image   (b) Grey image   (c) Resized image (d) Normalized modal image   (e) Diagonalization image

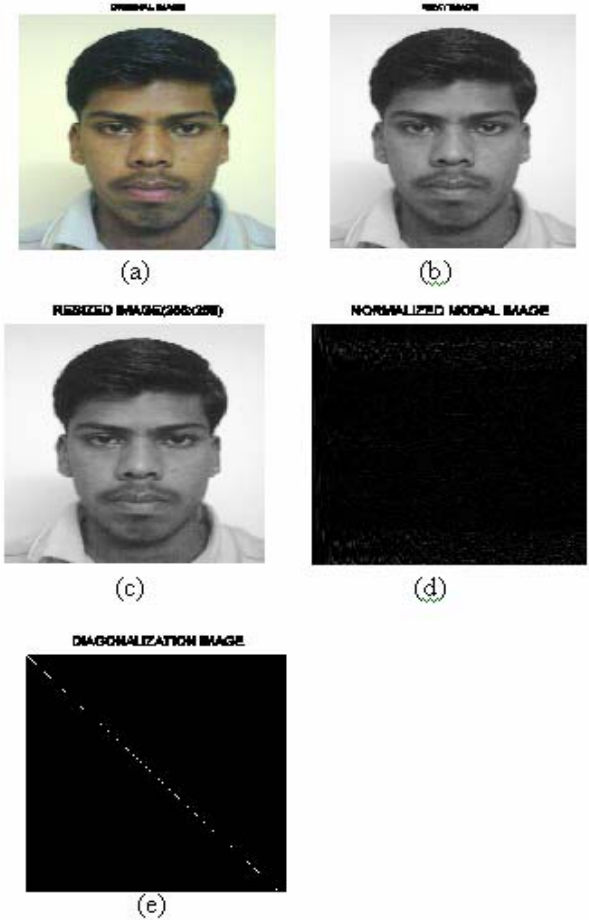

Fig.3. Canonical form based Face images.
(a) Original image   (b) Grey image   (c) Resized image
(d) Normalized modal image   (e) Diagonalization image

The multimodal system has been designed at multi-classifier & multimodal level. At multi-classifier level, multiple algorithms are combined better results. At first experimental the individual systems were developed and tested for FAR, FRR & accuracy. Table1 shows FAR, FRR & Accuracy of the systems.

Table1: The Accuracy, FAR, FRR of face & palmprint

| Trait | Algorithm | FAR | FRR | Accuracy |
|---|---|---|---|---|
| Face | Canonical form based | 4.5% | 8.7% | 97% |
| Palmprint | | 1.5% | 2.0% | 96% |

In the last experiment both the traits are combined at matching score level using sum of score technique. The results are found to be very encouraging and promoting for the research in this field. The overall accuracy of the system is more than 97%, FAR & FRR of 2.4% & 0.8% respectively.

## 6 Conclusion

Biometric systems are widely used to overcome the traditional methods of authentication. But the unimodal biometric system fails in case of biometric data for particular trait. Thus the individual score of two traits (face & palmprint) are combined at classifier level and trait level to develop a multimodal biometric system. The performance table shows that multimodal system performs better as compared to unimodal biometrics with accuracy of more than 98%.

## References

[1] Ross.A.A, Nandakumar.K, Jain.A.K. Handbook of Multibiometrics. Springer-Verlag, 2006.
[2] Kumar.A, Zhang.D Integrating palmprint with face for user authentication. InProc.Multi Modal User Authentication Workshop, pages 107–112, 2003.
[3] Feng.G, Dong.K, Hu.D, Zhang.D When Faces Are Combined with Palmprints: A Novel Biometric Fusion Strategy. In Proceedings of ICBA, pages 701–707, 2004.
[4] G. Feng, K. Dong, D. Hu & D. Zhang, when Faces are combined with Palmprints: A Noval Biometric Fusion Strategy, ICBA, pp.701-707, 2004.
[5] M. Turk and A. Pentland, "Face Recognition using Eigenfaces", in Proceeding of International Conference on Pattern Recognition, pp. 591-1991.
[6] M. Turk and A. Pentland, "Face Recognition using Eigenfaces", Journals of Cognitive Neuroscience, March 1991.
[7] L. Sirovitch and M. Kirby, "Low-dimensional Procedure for the Characterization of Human Faces", Journals of the Optical Society of America, vol.4, pp. 519-524, March 1987.
[8] Kirby.M, Sirovitch.L. "Application of the Karhunen-Loeve Procedure for the Characterization of Human Faces", IEEE Transaction on Pattern Analysis and Machine Intelligence, vol. 12, pp. 103-108, January 1990.
[9] Daugman.J.G, "High Confidence Visual Recognition of Persons by a Test of Statistical Independence", IEEE Trans. Pattern Analysis and Machine Intelligence, vol. 15, no. 11, pp. 1148-1161, Nov. 1993.

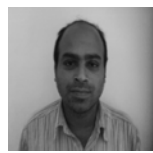

**Nageshkumar M.**, graduated in Electronics and Communication from Mysore University in 2003, received the M-Tech degree in Computer Science & Engineering from V.T.U., Belguam, presently pursing Ph.D under Mysore University. He was lecturer in J.V.I.T.

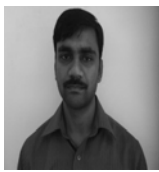
**Mahesh. P.K.**, graduated in Electronics and Communication from Bangalore University in 2000, received the M-Tech degree in VLSI design & Embedded Systems from VTU Belguam, presently pursing Ph.D under Mysore University. He was lecturer in J.S.S.A.T.E., Nodia and later Asst. Professor in J.V.I.T.

**Dr. M.N.Shanmukha Swamy**, graduated in Electronics and Communication from Mysore University in 1978, received the M-Tech degree in Industrial Electronics from Mysore University and then received PhD from Indian Institute of Science, Bangalore. Presently he is working as a Professor in S.J.C.E., Mysore. So for he has more than 10 research papers published, journals, articles, books and conference paper publications.